\documentclass[final,5p,times,twocolumn]{elsarticle}

\usepackage{graphicx}
\usepackage{multirow}
\usepackage{longtable}
\usepackage{rotating}
\usepackage{color}
\usepackage{makecell}
\usepackage{hyperref}
\usepackage{cite}
\usepackage{lineno}
\usepackage{amsfonts}
\usepackage{amsmath}
\usepackage{stfloats}
\usepackage{booktabs}
\usepackage[table,xcdraw]{xcolor}
\usepackage{algorithm}
\usepackage{algorithmicx}
\usepackage{algpseudocode}
\usepackage{float}
\usepackage{ulem}
\usepackage{soul}
\soulregister\cite7 
\soulregister\citep7 
\soulregister\citet7 
\soulregister\ref7
\soulregister\pageref7

\graphicspath{{figures/}}

\journal{Information Fusion}

\begin{document}

\begin{frontmatter}



\title{PPT Fusion: Pyramid Patch Transformer \\for a Case Study in Image Fusion}


\author[a]{Yu~Fu}
\author[a]{Tianyang~Xu}
\author[a]{Xiao-Jun~Wu\corref{correspondingauthor}}

\author[b]{Josef~Kittler}

\address[a]{Jiangsu Provincial Engineering Laboratory of Pattern Recognition and Computational Intelligence, \\ School of Artificial Intelligence and Computer Science, Jiangnan University, \\ 214122, Wuxi, China \fnref{address1}}

\address[b]{Centre for Vision, Speech and Signal Processing, University of Surrey, Guildford, GU2 7XH, UK \fnref{address2}}

\cortext[correspondingauthor]{Corresponding author email: wu\_xiaojun@jiangnan.edu.cn, other author email: yu\_fu\_stu@outlook.com,tianyang\_xu@163.com}

\begin{abstract}
The Transformer architecture has witnessed a rapid development in recent years, outperforming the CNN architectures in many computer vision tasks, as exemplified by the Vision Transformers (ViT) for image classification. 
% However, most of the existing visual Transformer models focus on the global correlation among the entire image cells, without explicitly constrain the local features like in CNN, such as texture and edges. 
However, existing visual transformer models aim to extract semantic information for high-level tasks, such as classification and detection.
These methods ignore %distorting 
the importance of the spatial resolution of the input image, thus sacrificing 
the local correlation information of neighboring pixels.
%the capacity in reconstructing the input or generating high-resolution images. 
% The Transformer models cannot keep the resolution of the features unchanged.
% Therefore, the features cannot reconstruct the original image, nor generate an effective high-resolution image. 
% These features lose most of the detailed information. 
% The features cannot reconstruct the original image, nor generate an effective high-resolution image. 
In this paper, we propose a Patch Pyramid Transformer(PPT) to effectively address the above issues.
Specifically, we first design a Patch Transformer to transform the image into a sequence of patches, where transformer encoding is performed for each patch to extract local representations. 
In addition, we construct a Pyramid Transformer to effectively extract the non-local information from the entire image. 
After obtaining a set of multi-scale, multi-dimensional, and multi-angle features of the original image, we design the image reconstruction network to ensure that the features can be reconstructed into the original input. 
To validate the effectiveness, we apply the proposed Patch Pyramid Transformer to image fusion tasks. 
The experimental results demonstrate its superior performance, compared to the state-of-the-art fusion approaches, achieving the best results on several evaluation indicators. 
Thanks to the underlying representational capacity of the PPT network, it
%is reflected by its universal power in feature extraction and image reconstruction, which 
can directly be applied to different image fusion tasks without redesigning or retraining the network.

\end{abstract}


\end{frontmatter}


\section{Introduction}
\label{intro}
In recent years, 
%Self-attention 
Transformer models have received wide attention from the research community recently, thanks to their promising performance in many visual tasks.
% \emph{\fu{(I replaced "self-attention" with "transformer", so there is no need to explain "The connection between self-attention and transformer". Anyway, the focus of the article is "transformer" rather than "self-attention".)}}
Vaswani et al. proposed the Vision Transformer structure at NIPS 2017, as a 2D data extension of the transformer originally designed for the Natural Language Processing (NLP) 
% \xu{(Unfold abbreviations the first time they appear in one paper.)}
tasks, such as Bert \citep{devlin2018bert} using Transformer as encoder, GPT \citep{radford2019language} using Transformer as decoder, Transformer-XL \citep{dai2019transformer} solving the problem of long sequences, etc. 
%\xu{(1. The connection between self-attention and transformer is missing in the above paragraph. 2. The underlying significance of studying the self-attention is not discussed, which should be essential in the first paragraph.)}
% They all use the transformer structure as their backbone.
It is also widely used in the field of Recommendation Systems to improve their performance. Examples include BST \citep{chen2019behavior} for behavioral sequence modeling, Autoint \citep{song2019autoint} for feature combination of CTR(Click-Through-Rate) prediction model, re-ranking model PRM \citep{pei2019personalized}, etc. 
Recently in the field of computer vision, many excellent methods demonstrated that Transformer can also obtain promising performance in tasks, such as image classification ViT \citep{dosovitskiy2020image}, object detection DETR \citep{zhu2020deformable}, semantic segmentation SETR \citep{zheng2020rethinking}, 3D point cloud processing Point Transformer \citep{pan20203d}, image generation TransGAN \citep{jiang2021transgan}, etc. 

Generally speaking, the transformer uses the self-attention module, which models global relations between patches for encoding features. 
The global operations provide a view of the input image which is more complete than local convolutions for semantic tasks such as classification, detection, and recognition. 
But for other visual tasks such as image fusion, defogging, super-resolution reconstruction, etc., such global aggregations would sacrifice 
%the reconstruction capacity, 
the local feature extraction capacity, 
degrading its effectiveness in low-level vision fields.
Therefore, in this paper, we explore the potential of Transformer in low-level vision tasks, balancing its semantic feature-saliency perception and pixel-level perception.
%xu{(1. The above two paragraph should be merged together. 2. Besides its popularity, more discussion should be added to involve its intrinsic merit.)}

In principle, the seminal work on ViT proves that using a pure Transformer network could achieve the state-of-the-art performance in image classification.
%\xu{('SOTA' is not a optimal expression for a journal paper.)}
Specifically, ViT splits the input into 14$\times$14 or 16$\times$16 patches.
Each patch is flattened to a vector, acting as a token for the Transformer system.
% ViT inputs these tokens into the Transformer model, just like the NLP task, learning the global relationship.
% And finally using these features for image classification.
ViT is the first transformer model designed to extract image features, but it still suffers the following two limitations:
1) ViT requires a huge dataset such as the JFT-300M dataset for pre-training to better explore the relationship between pixels. 
It cannot obtain satisfactory results using a midsize dataset such as ImageNet-1K.
2) The Transformer structure in ViT extracts non-local interactions from the entire image, which is not suitable for learning local patterns, including texture and edges.
We believe that local patterns are essential for visual recognition tasks, which has been proved by existing CNN(Convelutional Neural Network) techniques.
In particular, CNN extract various shallow features, and then obtains semantic information through multi-layer nonlinear stacking. 
% The global information extracted by ViT may be sufficient for image classification or detection. 
%Therefore, features obtained by ViT are difficult to generate high-resolution images, impeding its reconstruction capacity in many low-level visual tasks.
Therefore, ViT focus on the semantic features while ignoring shallow features, which is not conducive to low-level visual tasks.
%{\color{red}3) A standard Transformer calculates the relationship between all tokens, that is, each pixel .It will resulting in excessive computing and storage consumption. ViT cleverly splits the image into several patches, and identify each patch as a new computing element. If the original image size is 256$\times$256 and the 16$\times$16 patch is used for split, the 16$\times$16 features will be obtained. We simply consider that to some extent, the image is down-sampled 16 times non-linearly. Although the semantic features can be learned, the low-level features cannot be restored.
	%}
Motivated by this analysis, we design a new transformer model to address the above issues, endowing the Transformer with pixel-level perception.

To extract low-level visual representations, we divide the original image into several patches of appropriate size.
For each patch, the transformer is used to calculate the global correlation features of all pixels, with which the source image patches can be reconstructed. 
% and these features are finally reconstructed into the same size of the source image. 
Therefore, we can obtain a set of low-level features of the image, sharing similar characteristics.
% Each patch gathers the features of all pixels in the patch, but there is a lack of correlation calculation between patches. 
We coin this module as "Patch Transformer". 
Different from ViT, as shown in Fig. \ref{fig:patch}, Patch Transformer operate on all pixels in each patch, 
%without performing dimensional compression, thus preserving the pixel-level clues for reconstructing the image.
focusing on extracting the correlation of local pixels, thus extracting more local features.
Second, to explicitly extract the global patterns of the image, we design a down-sampling pyramid to achieve local-to-global perception.
Specifically, we down-sample the input image once, obtaining a half size image, where we perform Patch Transformer.
% Now each patch of the down-sampled image features corresponds to four adjacent blocks in the original image.
% The four adjacent blocks are also gathered into a certain patch feature. 
After repeating down-sampling and performing the Patch Transformer until the image becomes the same size as the predefined patch.
The obtained features corresponding to different scales  are then up-sampled to the original size. 
% We get a set of different scales features including local information and global information.
We name this operation as “Pyramid Transformer”.
% \xu{(as these two modules are the most essential contributions of this paper, so more discussions should be involved in this paragraph, highlighting their differences)}
% The differences are discussed in detail in related work. Do I need a long discussion here? I am worried about the repetition of the description. I don't understand how to modify this part. Can I move some discussions below here?)}}
Based on the Patch Transformer and Pyramid Transformer, we develop a Pyramid Patch Transformer (PPT) image feature extraction model. 
The PPT model can fully extract the features of the input image, including local context and global saliency. 
We further design a reconstruction auto-encoder network to ensure that the extracted features can reconstruct the input image.

\begin{figure}[!ht]
\centering
\includegraphics[width=\linewidth]{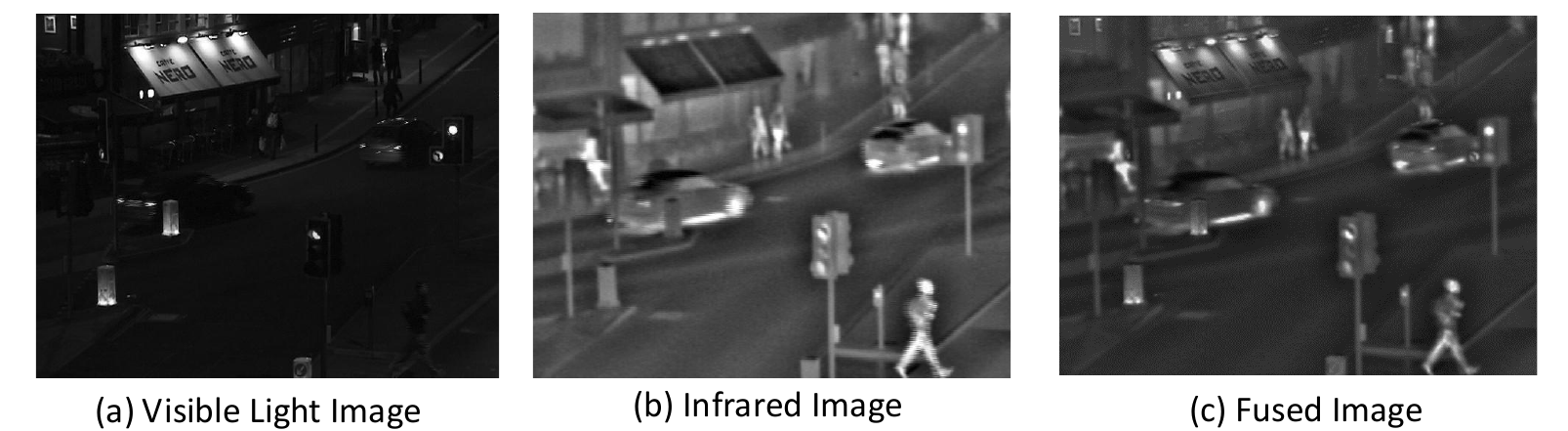}
\caption{Image Fusion. (a) is a visible light image. (b) is an infrared image. (c) is a fused image.}
\label{fig:fuseintro}
\end{figure}

To verify the effectiveness of the proposed PPT model, we apply this feature extraction approach to the image fusion task. 
The input images are obtained by different kinds of sensors, such as infrared images and visible light images, medical CT images and X-ray images, or images with different focus, images with different exposures and so on. 
These multi-source images reflect different physical attributes of the same scene. 
In general, image fusion tasks aim to fuse images of different modalities according to a certain algorithm, so that the fused image contains more information than any single image. 
An example of the fusing a visible light image and an infrared image is shown in Fig.\ref{fig:fuseintro}. 
The fused images contain more comprehensive information for easy recognition and observation by humans and machines.

Overall, the proposed Pyramid Patch Transformer possesses the following merits.
First of all, compared to ViT that only focuses semantic information, Patch Transformer can extract low-level texture features of the image. 
Secondly, compared with CNN, PPT uses Pyramid Transformer to calculate global correlation, its receptive field is the entire image.
Besides, the common Transformer model requires a large data set, and we only need to train PPT on MS-COCO \citep{lin2014microsoft} and Imagenet \citep{deng2009imagenet} with less computing resources.
Finally, to the best of our knowledge, we are the first to perform image fusion tasks using an effective low-level vision Transformer model, which is entirely convolution-free. 
Compared with the existing image fusion methods, the fusion image of PPT have achieved promising results in both subjective and objective evaluation.

The PPT model can extract a variety of multi-scale features from the input image.
We use PPT model as the feature extraction network, and then design a feature decoder for feature compression and image reconstruction. 
% Based on the autoencoder paradigm, we use PPT model to extract the features of the input image, and then use the designed fusion strategy to fuse the features to obtain a set of fused features. 
% Finally we input the fused features into the feature decoder to obtain the fused image.
In summary, our contributions are four-fold:

%	\begin{itemize}
%		\item An improved Patch Transformer to extract low-level image representations without loss of resolution, integrating the interactions among raw pixels.
%
%		\item A novel Pyramid Transformer to reflect the global relationships of multi-scale patches, achieving local-to-global perception.
%
%		\item A new Pyramid Patch Transformer as a general feature extraction module, which is successfully applied to image fusion tasks with superior performance against the state-of-the-art methods.
%	\end{itemize}

\begin{itemize}
\item A new Pyramid Patch Transformer as a general feature extraction module for low-level vision tasks is proposed, supplementing the 
%reconstruction 
local feature extraction
capacity of transformer in the CV field, with its successful application in the field of image fusion, demonstrating its feasibility.

\item The designed Patch Transformer can reflect intrinsic local features via directly modelling the pixels in each Patch. Besides, the proposed position embedded dimension extends to all spatial pixels, achieving fine-grained location-aware attention against the other Transformers.

\item A novel Pyramid Transformer is constructed to reflect the global relationships of multi-scale patches, achieving local-to-global perception.

\item The receptive field is applied to the Transformer model for the first time. In particular, existing Transformers can only aggregate global features. But our PPT can achieve multi-range and multi-level pixel perception like CNN, hierarchically passing the visual clues.
\end{itemize}

\section{Related Work}
\subsection{Transformer in Vision}
Transformer is a powerful model in the NLP field. 
It has swept all NLP areas, with the eye-catching title and excellent performance since \textit{Attention is All You Need}. 
Recently, Transformer has extended in the CV field. 
For example, a self-supervised learning method with Vision Transformers as the backbone architecture MoB \citep{xie2021self}, a new tracker network based on powerful attention mechanism TrTr \citep{zhao2021trtr}, a deformation field learning method that solves the problem of deformable image registration AiR \citep{wang2021attention}, a new Semantic segmentation method Segmenter \citep{strudel2021segmenter}, and more excellent Transformer models for other vision tasks such as image classification \citep{chen2020generative,dosovitskiy2020image,wu2020visual}, object detection \citep{carion2020end,zhu2020deformable,amato2019learning}, etc.
Among these works, most of them follow the Vision Transformer (ViT) to establish the corresponding models. 

Typically, a Transformer encoder is composed of a multi-head self-attention layer (MSA) and a Multi-layer Perception (MLP) block. 
A Layer Norm (LN) operation is used, with the residual structure, before each MSA and MLP layer.
The essential design of Transformer is that the input vector is combined with the position embedding to preserve the localization clues for each token.
To introduce the basic Transformer to the visual task, 
%	ViT splits the input image $\mathbf{x}\in \mathbb{R}^{H\times W\times C  }$ into a sequence of several patches  $\mathbf{x}_p\in \mathbb{R}^{\left(P^{2} \cdot C\right) \times N}$, where $H,W,C$ are the corresponding resolution and number of channels, $P$ is the width / height of the patch, and $N=HW/P^2$. 
%	ViT maps these patches to the $D$ dimensional features after forward passing the network. 
ViT splits an input image $X\in \mathbb{R}^{H\times W }$ into $w\times  h$ patches by a $p\times p$ size sliding window to obtain a 1-D vector $Y1\in \mathbb{R}^{h\times w\times p^2}$.
However, ViT maps $Y1$ to $E\in \mathbb{R}^{p^2\cdot C\times D}$, which ignores the fine-grained information of all pixels in the $ w\times  h$ patch, only retaining the semantic information between patches. 
ViT uses 16 vectors represents the $w \times h$ size patch feature. 
The obtained output at $\mathbf{z}_{L}^{0}$ is the classification result. 
A ViT network structure for image classification is as follows,
\begin{equation}
	\left\{
	\begin{aligned}
		\mathbf{z}_{0} &=\left[\mathbf{x}_{\text {class }} ; \mathbf{x}_{p}^{1} \mathbf{E} ; \mathbf{x}_{p}^{2} \mathbf{E} ; \cdots ; \mathbf{x}_{p}^{N} \mathbf{E}\right]+\mathbf{E}_{p o s}, \\& \mathbf{E} \in \mathbb{R}^{\left(P^{2} \cdot C\right) \times D}, \mathbf{E}_{p o s} \in \mathbb{R}^{(N+1) \times D} \\
		\mathbf{z}_{\ell}^{\prime} &=\operatorname{MSA}\left(\operatorname{LN}\left(\mathbf{z}_{\ell-1}\right)\right)+\mathbf{z}_{\ell-1}, \ \ \ \  \ell=1 \ldots L \\
		\mathbf{z}_{\ell} &=\operatorname{MLP}\left(\operatorname{LN}\left(\mathbf{z}_{\ell}^{\prime}\right)\right)+\mathbf{z}_{\ell}^{\prime}, \ \ \ \  \ell=1 \ldots L \\
		\mathbf{y} &=\operatorname{LN}\left(\mathbf{z}_{L}^{0}\right) 
	\end{aligned}
	\right..
\end{equation}

\begin{figure*}[t]
\centering
\includegraphics[width=\linewidth]{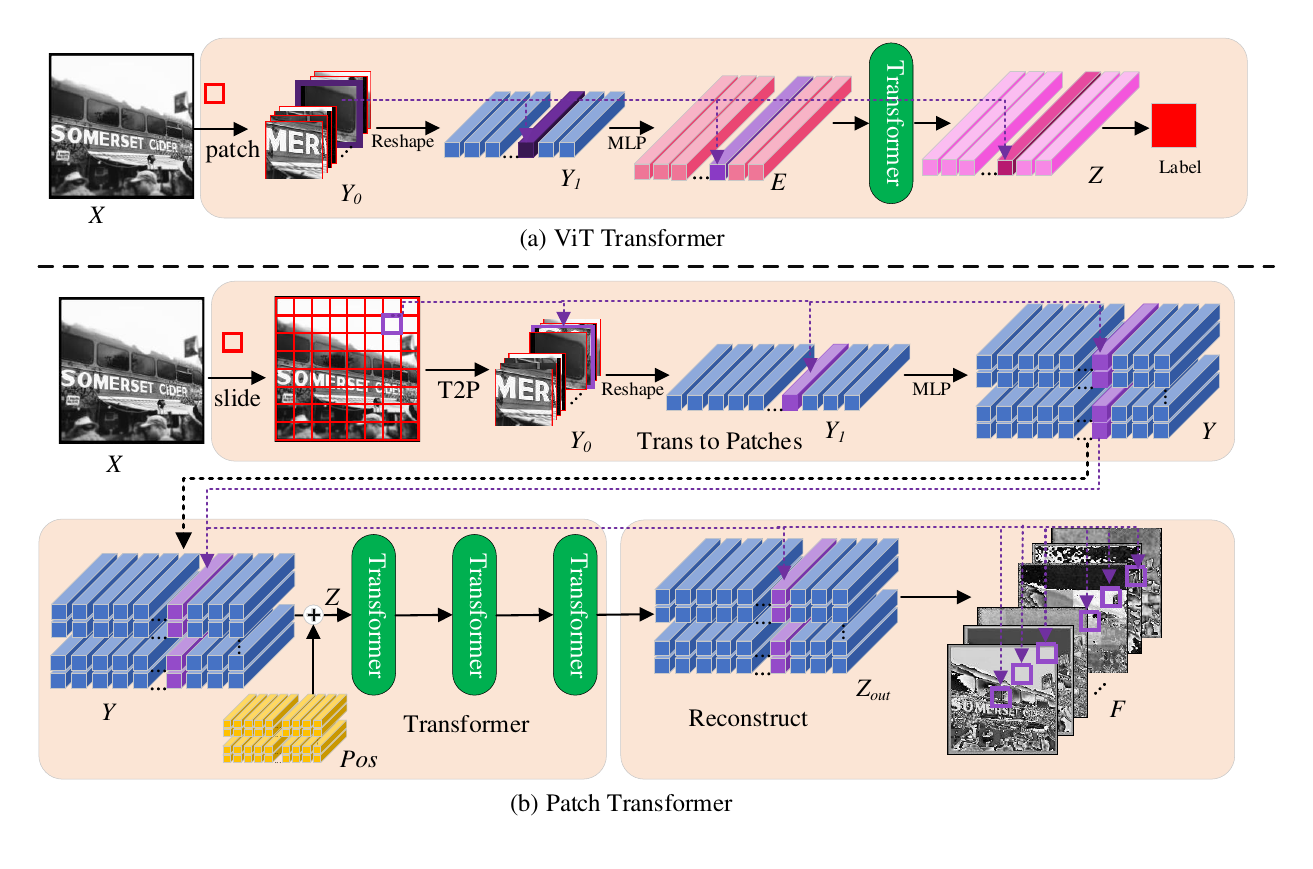}
\caption{Comparison between ViT and Patch Transformer. (a) Illustration of the ViT. ViT maps the image to a high-latitude space, learning more semantic information but also destroying the pixel-level relationship.
	(b) Illustration of the Patch Transformer. The first step is to split the input to multiple patches. A $p\times p$ sliding window is used to split the image into a sequence of patches. Each patch is reshaped into a 1-D vector, and an MLP is used to extend the channels. The second step is to use the multi-layer Transformer module to obtain the non-local representations within this patch. The third step is to reconstruct the two-dimensional patch from the learned representations.}
\label{fig:patch}
\end{figure*}

% As the title of ViT, "An Image is Worth 16x16 Words". 
ViT  roughly split images as a $16 \times 16$ size vector. 
It is useful for semantic information extraction, but it is disastrous for 
%fine-grained pixel retention 
local features
of low-level visual tasks.
%To a certain extent, we regard ViT as a special complex image downsample operation which highlights semantic information. 
In Fig. \ref{fig:patch}(a), we simply draw the ViT process.
Purple highlight the process of obtaining a patch token.

In addition, SiT\citep{atito2021sit} supplements the lack of image reconstruction tasks in self-supervised low-level vision. 
Sit perform well in tasks including image reconstruction, rotation prediction and contrastive learning. 
SiT uses the global computing power of Transformer to model semantic concepts within patches. 
Similarly, it does not attach importance to shallow information.

To summarize, most existing Transformer model also focuses on solving high-level semantic tasks, while there are still few explorations for low-level visual tasks.

\subsection{Image Fusion}
% \emph{\fu{(I cut the “image fusion introduction” here, as "the development of image fusion", the following is a brief formula to the existing methods, and then a brief analysis of the shortcomings of the existing image fusion methods.)}}
There are numerous algorithms that can achieve promising image fusion performance with the developing computer vsion techniques \citep{li2017pixel,ma2019infrared}. 
Existing image fusion approaches can be roughly divided into two categories: traditional methods and CNN-based methods.

Traditional image fusion methods use algorithms to decompose or convert images to other domains for fusion, and then inversely transform them into general domain images. 
In this category, 1) use pyramid \citep{mertens2009exposure}, curvelet \citep{zhang1999categorization}, contourlet \citep{upla2014edge}, etc. to perform multi-scale image decomposition. 
2) use Sparse Representation (SR) \citep{zong2017medical}, Joint Sparse Representation (JSR) \citep{zhang2013dictionary} or approximate sparse representation \citep{bin2016efficient} to represent the images in the sparse subspace. 
3) perform low-rank representation of the image such as LRR \citep{li2017multi} or MDLatLRR \citep{li2020mdlatlrr}. 
4) convert images to a subspace domain, such as PCA \citep{pca2017}, ICA \citep{ICA2007}, NMF \citep{mou2013image}. 

Image fusion approach based on CNN can be divided into two categories: 1) The method based on autoencoders uses an encoder to extract features into the latent space for feature fusion, and then the fused features are input to the decoder to obtain the fused image \citep{prabhakar2017deepfuse,li2018densefuse,li2020nestfuse,fu2021dual} . 2)The end-to-end fusion network designs a suitable structure and loss function to realize the end-to-end image generation  \citep{ma2019fusiongan,ma2020pan,fu2021image}.
In ICCV 2017, Prabhakar et al. proposed DeepFuse \citep{prabhakar2017deepfuse} approach, introducing the auto-encoder structure for multi-exposure image fusion task. 
% {\color{red} This is a great innovation in image fusion using anto-encoder.}
Specifically, DeepFuse trains the auto-encoder network, with the encoder extracting features of the image.
After performing an addition fusion strategy in the middle layer, the fused features are input to the decoder to obtain the final output. 
Similar structures are further developed by DenseFuse \citep{li2018densefuse} and IFCNN\citep{zhang2020ifcnn}. 
% Besides, IFCNN defines a auto-encoder fusion framework for multiple fusion tasks.
The general steps of these auto-encoder-based image fusion methods are as follows: 
\begin{equation}
	\left\{
	\begin{aligned}
		&X_{features} = \operatorname{Encode}(X_{input})\\
		&Y_{features} = \operatorname{Encode}(Y_{input})\\
		&Z_{features} = \operatorname{Fusion} (X_{features},  Y_{features}) \\
		&Z_{output} = \operatorname{Decode}(Z_{features})
	\end{aligned}
	\right..
\end{equation}

The learned convolution kernels can gradually extract high-level abstract features.
In theory, the receptive field should cover the entire image, but \citep{2017Understanding} have shown that the actual receptive field is much smaller than the theoretical one.
This impedes the full use of context information to capture the visual features.
Although we can continuously deepen the convolutional layers, this would obviously heavy the model and increase the amount of calculation dramatically.
Therefore, the CNN based deep learning image fusion algorithm cannot support a suitable scope for perceiving the visual clues. 
After analysing the deep learning auto-encoder image fusion framework, we conduct an exploration to to use Pyramid Patch Transformer instead of the traditional CNN to address the above limitations.

%
%\begin{figure*}
%\begin{center}
%\fbox{\rule{0pt}{2in} \rule{.9\linewidth}{0pt}}
%\end{center}
%   \caption{Example of a short caption, which should be centered.}
%\label{fig:short}
% ;\end{figure*}
\section{Pyramid Patch Transformer}
For high-resolution images, most of the existing transformer approaches split the image into several patches.
% , and each patch is used as a vector.
Suppose the resolution of an image is $W\times H$, we split it into $w\times h$ vectors with each patch being $p\times p$, where $w=W/p$, $h=H/p$. 
% The input and output of the transformer are images with size $w\times h\times p^2$. 
Therefore, the features with size $w\times h$ are the 2D sequential representations of the original image $W\times H$ %after down-sampling and non-linear operators. 

Transformer then regards each patch as a minimum computing unit. It calculate the correlation between $w\times h$ patches. However, the $p\times p$ elements in the patches will be scrambled into the 2D vectors and will not directly participate in the correlation calculation which is named Attention in Transformer. This operation can strongly perceive the context information of any patch, but it is difficult to perceive the local information of pixels in the patch.

%However, the pixel-level original image information is mapped into a low-dimensional high-level feature space, with semantic meaning and discrimination. 
% However, if this group of low-dimensional features are reconstructed, 
%It is difficult to reconstruct the original image with the obtained Transformer features.

To reflect the pixel details, a straightforward solution is to set the size of the patch to 1. 
This means that the Transformer is performed on the original resolution image, resulting in inoperable issue. 
For example, if a transformer is applied to a $256\times256$ image, at least one attention matrix of $(256\times256)*(256\times256)=4294967296$ parameter will be generated, which requires huge memory.
In order to overcome this problem, we propose the Pyramid Patch Transformer, a network framework that uses fully-transformer for image feature extraction.
The PPT model consists of two parts: 1). a memory-friendly Patch Transformer for large-resolution images. 2). Pyramid Transformer designed to capture the global information of the local patches. 

PPT model can perceive the local information of pixels in any patch and the contextual information between patches.
% We will now explain these two modules one by one.

\subsection{Patch Transformer}
The Patch Transformer module is designed to alleviate the  memory consumption caused by general transformers with exceeding tokens when processing large-resolution images. 
Each Patch Transformer module contains three steps: 1, Trans to Patches, 2, Transformer and 3, Reconstruct. The Patch Transformer process is shown in the Fig.

Patch Transformer performs weight-sharing Transformers for each patch to calculate the correlation of all pixels in the patches and obtain local information in each patch of the input image.

\begin{figure*}[!ht]
	\centering
	\includegraphics[width=13cm]{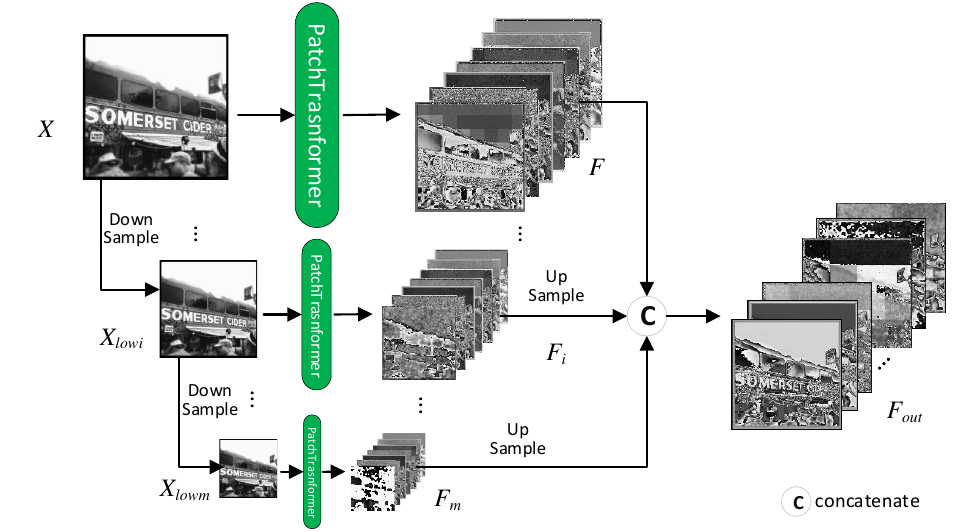}
	\caption{Pyramid Transformer. The input image $X$ is down-sampled $m$ times. Each time the down-sampled image employs the Patch Transformer to extract features to obtain $F_i$. Finally, all different scaled features $F_i$ are upsampled and concatenated to obtain $F_{out}$.}
	\label{fig:pyramid}
\end{figure*}

\begin{figure}[!ht]
	\centering
	\includegraphics[width=7cm]{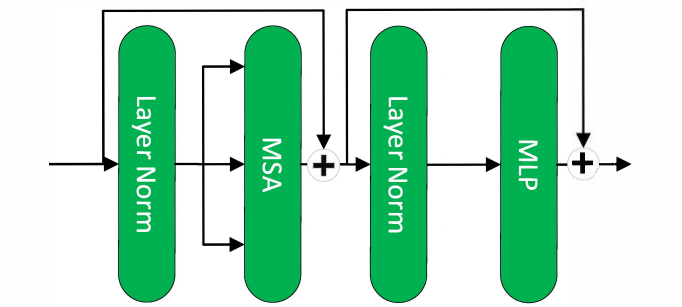}
	\caption{Transformer module.}
	\label{fig:transformerl}
\end{figure}

%\subsubsection{Trans to Patches}
As shown in the Fig. \ref{fig:patch}(b), given an $H\times W$ input image $X\in \mathbb{R}^{H\times W }$,  $X$ is split into $w\times h$ patches by a sliding window with $p\times p$ size, where $h=H/p, w=W/p$. 
We name this operation T2P. 
We can obtain a sequence of $h\times w\times p\times p$ patch features $Y_0$, $ Y_0=[{x_0, x_1, x_2\cdots x_{h\times w} }]$, $x_i \in \mathbb{R}^{p\times p }$. 
We reshape each $x_i$ into a 1-D vector. 
$Y_0\in \mathbb{R}^{h\times w\times p\times p}$ is reshaped into $Y_1 \in \mathbb{R}^{h\times w\times p^2}$.
In order to enhance the information, we use MLP to increase the dimension of $Y_1$ to obtain $Y\in \mathbb{R}^{h\times w\times p^2\times C}$, equivalent to a sequence of $h\times w$ patch features with size with $p^2$ and $C$ channels, $ Y=[{y_0, y_1, y_2\cdots y_{h\times w} }]$, $y_i \in \mathbb{R}^{p^2\times C }$.
\begin{equation}
	\left\{
	\begin{aligned}
		&Y_0 = \operatorname{T2P}( X )\\
		&Y_1 = \operatorname{Reshape}( Y_0 )\\
		&Y = \operatorname{MLP}(Y_1)
	\end{aligned}
	\right..
\end{equation}
We set a learnable position embedding vector $Pos$. 
In addition, we extend the vector $Pos$ to the same dimension as $Y$, $Pos\in \mathbb{R}^{h\times w\times p^2\times C}$, enabling learning the location clues among the embedding vectors. 
% We embed the location information $Pos$ in the Patch feature $Y$ to
Therefore, we can obtain the feature $Z\in \mathbb{R}^{h\times w\times p^2\times C}$.
\begin{equation}
	\begin{aligned}
		&Z = [y_0, y_1,\cdots y_{h\times w}]+Pos\\
	\end{aligned}
\end{equation}
Then we conduct transformers for each patch $z_i$ in $Z$, $Z = [{z_0, z_1, z_2\cdots z_{w\times h} }]$. 
The Transformer encoder module can be applied several times in the network. 
Each Transformer module is divided into two steps,multi-head self-attention layer (MSA) and Multi-layer Perception (MLP). 
A standard Transformer module with LayerNorm and Residual structure is shown in the Fig. \ref{fig:transformerl}.
\begin{equation}
	\left\{
	\begin{aligned}
		&z_i' = \operatorname{MSA}(\operatorname{LN}(z_i))+z_i\\
		&z_i'' = \operatorname{MLP}(\operatorname{LN}(z_i'))+z_i'
	\end{aligned}
	\right..
\end{equation}

We restore $Z_{out} = [{z''_0, z''_1, z''_2\cdots z''_{w\times h} }]$ according to the order of split $X$. 
The corresponding output is a set of features $F\in \mathbb{R}^{H\times W\times C} $ mapped from the original image to the latent space.
\begin{equation}
	\begin{aligned}
		F = \operatorname{Re}(Z_{out})\\
	\end{aligned}
\end{equation}

\subsection{Pyramid Transformer}
Pyramid Transformer uses the pyramid structure to perceive the contextual information between patches.
Using the above Patch Transformer, each input image will be split into several patches. 
% And we use Transformer module for each patch. 
% There is an obvious shortcoming. 
The representations of each patch are only related to the pixels within the patch, without considering the long-range dependency between pixels in the entire image.
% and the global feature is missing. 
To address this issue, we refer to the multi-scale approach with the following design to construct a pyramid structure.
First, the image is down-sampled once to obtain an image $X_{low}$ with size of $W/2 \times H/2$,  $X_{low}\in \mathbb{R}^{H/2\times W/2} $. 
Apply the corresponding Patch Transformer to $X_{low}$ to get the representations $F_{low}$, $F_{low}\in \mathbb{R}^{H/2\times W/2\times C} $. 
Then $F_{low}$ is upsampled to obtain $F_1$ with the same size as the input image $X$. 
\begin{equation}
	\left\{
	\begin{aligned}
		&X_{low}=\operatorname{Downsample}(X)\\
		&F_{low} = \operatorname{PatchTransformer}(X_{low})\\
		&F_{1} = \operatorname{Upsample}(F_{low})\\
	\end{aligned}
	\right..
\end{equation}

1) Continue to downsample the image $X_{low}$,

2) Use Patch Transformer to extract representations,

3) Upsample to $H\times W$ to get $F_i$.

Repeat the above operations recursively until the the downsampled image $X_{low}$ can be split into one patch.
We perform these operations $m$ times. 
Suppose the image is of $S\times S$ size, and the spitted patch is of $p\times p$ size, we can obtain $m = log_2(S/p)$.
After concatenating all the features at different scales, we get a set of multi-scale features $F_{out} = [{F, F_1, F_2\cdots F_m }] $, as shown in Fig.\ref{fig:pyramid}.
Patch Transformers in different pyramid scales focus on diverse granularities, such that different scales are specifically trained without sharing the weights. 
% Different scales features are in different dimensions, the mapping relations of PatchTransformers should be different.

\subsection{Transformer Receptive Field}
% We believe that Pyramid operation is feasible and reasonable. 
In general, CNN performs well in the computer vision field.
One essential factor is the receptive field of CNN, which can effectively capture the local features in each specific spatial scales of the image.
With the size of the convolution kernel increasing or the depth of the convolution layer deepening, each cell of the feature maps reflects a relevant spatial region in the original image.
\begin{figure}[!ht]
	\centering
	\includegraphics[width=5cm]{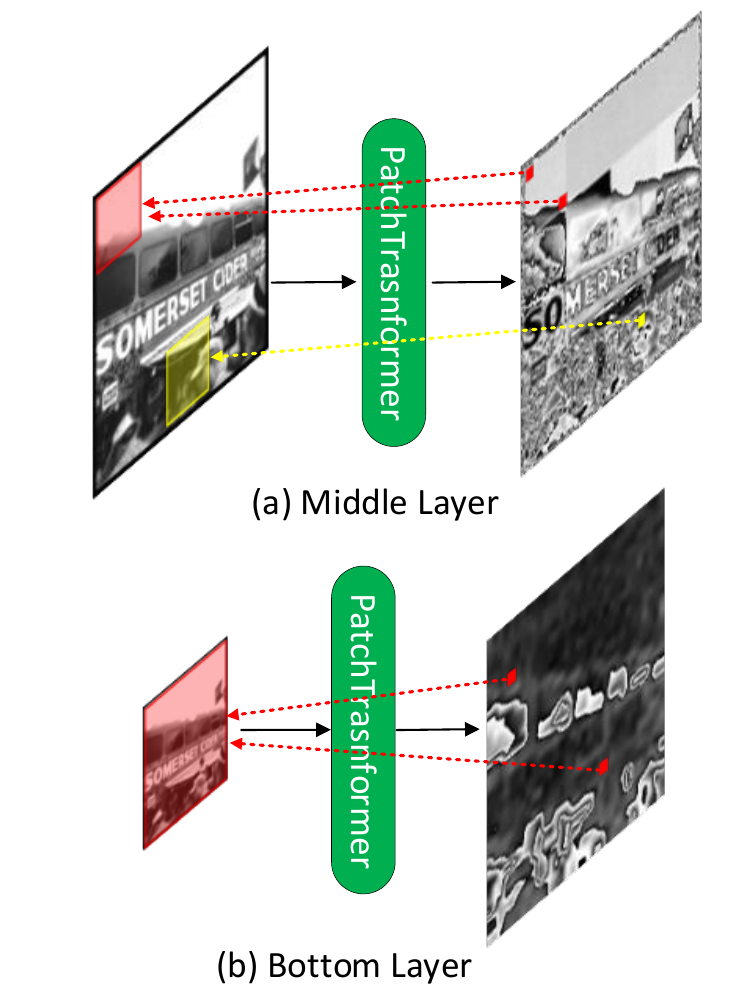}
	\caption{Transformer Receptive Field. a) is the mapping of the feature pixels in the middle of the Pyramid Transformer. b) shows the mapping in the bottom layer of the Pyramid Transformer. }
	\label{fig:receptive}
\end{figure}
As shown in the Fig. \ref{fig:receptive} (a), a Patch Transformer is performed on the image, with each pixel of its feature being associated with all the pixels of the entire patch. 
This patch size can be considered as the receptive field of this Patch Transformer.
The receptive field can expands four times larger after downsampling once, as the length and width of the image are both half.

With the gradual deepening of the pyramid, the range of associated pixels expands from the local area to global. 
The receptive field of the Patch Transformer has also become larger.
In particular, the pixels that are close to each other contribute more correlation, and the pixels with a long distance preserve weak long-range dependence.
With the bottom Patch Transformer layer in the Pyramid Transformer, the receptive field is expanded to the entire image, as shown in the Fig.\ref{fig:receptive} (b).
The continuous down-sampling is designed to obtain a large receptive field here. 
A large receptive field on the original image captures more large-scale or global semantic features with less detail information. 
% The lost detail information has little effect on semantic features. 
While the upper several layers in the Pyramid Transformer captures low-level details. 
Therefore, we argue that the Pyramid Transformer can extract both shallow and semantic information simultaneously.

\subsection{Network Architecture}
\begin{figure}[!ht]
	\centering
	\includegraphics[width=\linewidth]{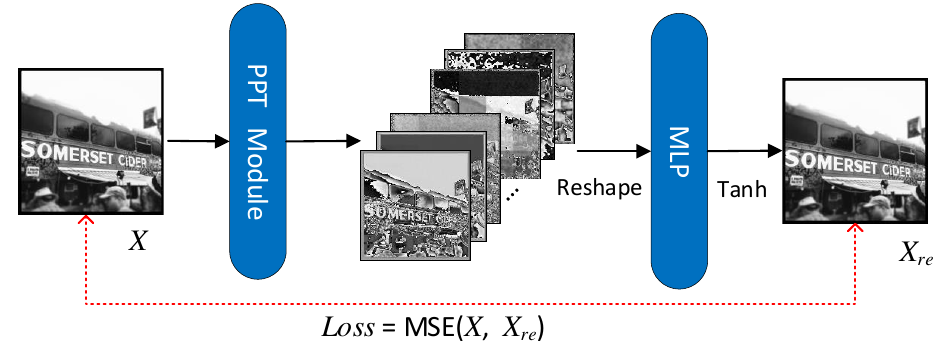}
	\caption{Network Architecture.}
	\label{fig:architecture}
\end{figure}

We design the auto-encoder network for image reconstruction, as shown in the Fig. \ref{fig:architecture}. 
% It can ensures that the features extracted by the PPT module are sufficiently complete.
The encoder is composed of the Pyramid Transformer and the Patch Transformer.
After obtaining a set of multi-scale features after encoding, the reconstructed image can be generated with the decoder. 

The decoder is an MLP composed of two Fully Connected (FC) layers, and uses the GELU \citep{hendrycks2016gaussian} activation function and Tanh activation to output.
\begin{equation}
	\begin{aligned}
		X_{re} = \operatorname{Tanh}(\operatorname{GELU}(\operatorname{FC}(\operatorname{GELU}(\operatorname{FC}(X)))
	\end{aligned}
	\label{equ:decoder}
\end{equation}

We use the mean square error (MSE) loss function as the reconstruction loss for the network.
\begin{equation}
	\left\{
	\begin{aligned}
		&Loss = \operatorname{MSE}(X,X_{re})\\
		&\operatorname{MSE}(X,Y)=\frac{1}{N}\sum_{n=1}^{N}(X_n - Y_n )^2 
	\end{aligned}
	\right..
\end{equation}

\subsection{Features Visualization}
\begin{figure}[!ht]
	\centering
	\includegraphics[width=\linewidth]{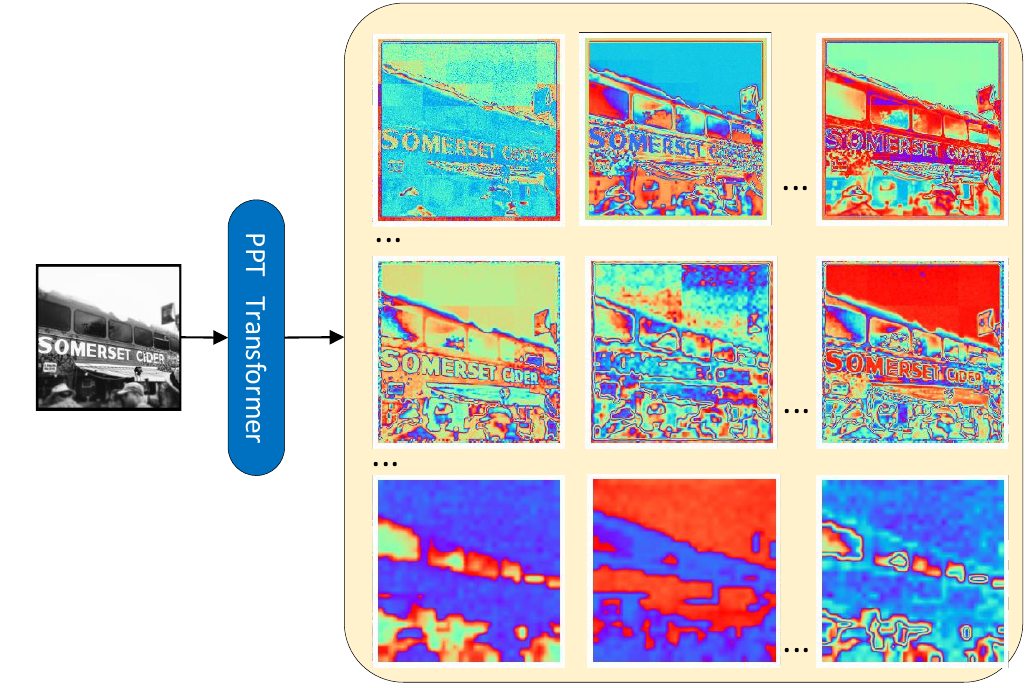}
	\caption{Features Visualization. Features in the first row are generated by the top layer in the Pyramid Transformer. They have the smallest receptive field and represent low-level features. Features in the third row are generated by the bottom layer of the Pyramid Transformer. They have the largest receptive fields and represent semantic features.}
	\label{fig:features}
\end{figure}

% We use the fully-transformer structure as the feature extractor and apply image reconstruction loss function. We believe that the obtained extracted features already contain almost the shallow features and semantic features of the source image.
As shown in the Fig. \ref{fig:features}, we obtain the extracted features by the PPT module. 
We select the features from three different receptive fields in the Pyramid Transformer. 
In the first row with the smallest receptive field, it can be seen that the features represent more low-level features such as edge contour and color distribution of the image.
While in the third row of the features with the largest receptive field, it can be seen that the features represent the concerned area of the related object, reflecting the semantic related feature of the pixels.

\section{Pyramid Patch Transformer For Image Fusion}
The primary purpose of the image fusion task is to generate a fusion image that contains as much useful information as possible from the two source images. 
We use the designed PPT module to extract the image features for image fusion tasks.

\begin{figure*}[!ht]
	\centering
	\includegraphics[width=\linewidth]{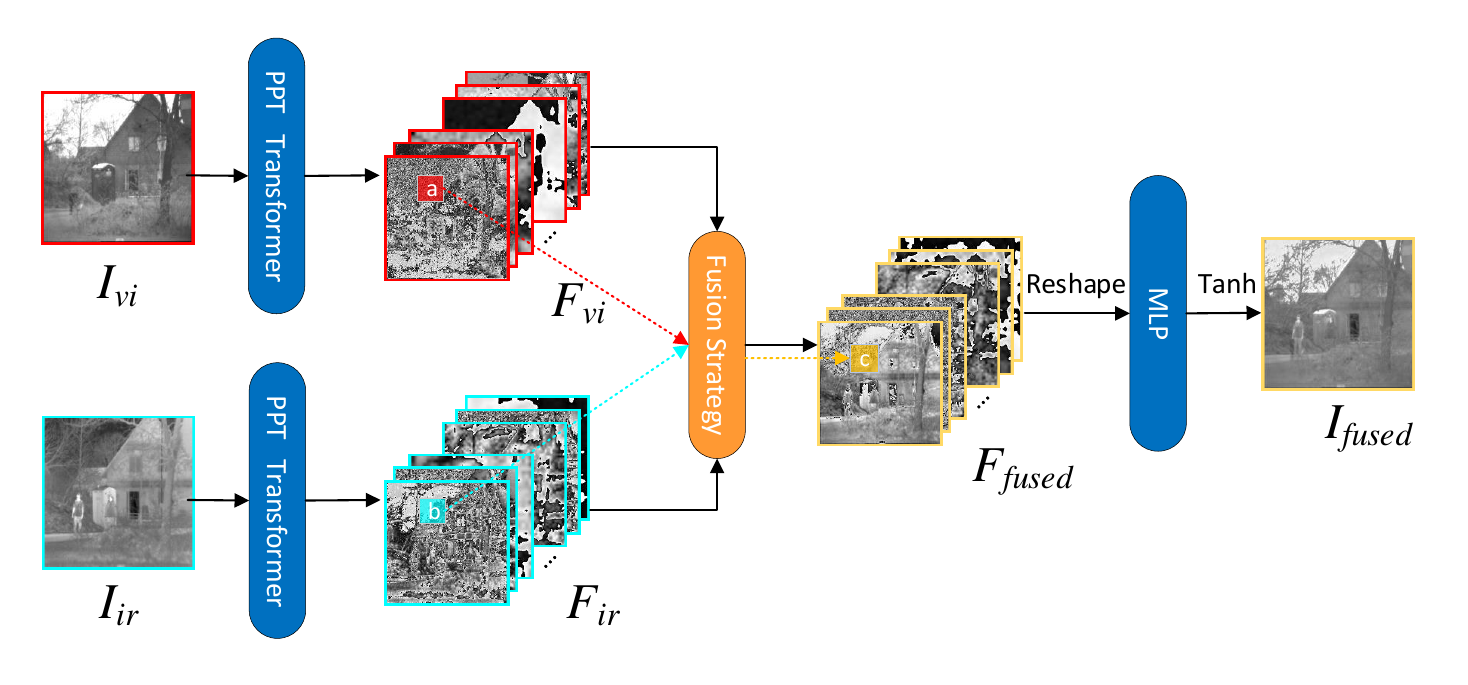}
	\caption{Image Fusion Network. Input the multi-source images to the PPT module to obtain the multi-source features. The fused features are decoded by MLP to obtain the fused image.}
	\label{fig:fusion}
\end{figure*}

\subsection{Fusion Network Architecture}
% When the PPT module is trained, it will be an good feature extractor. 
We take the infrared image and visible light image fusion task as an example. 
We input the visible light image $X_{vi}$ and the infrared image $X_{ir}$ into the pre-trained PPT encoder module to obtain $F_{vi}$ and $F_{ir}$.
The PPT encoder can map any image $X$ to a high-dimensional feature space to obtain features $F$. 
These features $F$ can represent the input image $X$ from different angles such as edge, texture, color distribution, semantic information, etc.. 
As we use the Siamese structure with a same PPT encoder module to extract features, $F_{vi}$ and $F_{ir}$ are mapped to the same feature space. 
We can easily perform fusion operations on $F_{vi}$ and $F_{ir}$ across the channel dimension, thus get a new fused feature representation $F_{fused}$.

Because the features $F$ obtained by the PPT encoder can be reversely mapped to the original space by the trained MLP decoder in Equ.\eqref{equ:decoder}, the $F_{fused}$ we calculated is also in the same feature space, we can reconstruct the fused features $F_{fused}$ to a fusion image $X_{fused}$ through the same MLP decoder, as show in Fig. \ref{fig:fusion}.

\subsection{Fusion Strategy}
For different image fusion tasks, we choose different fusion strategies. 
All fusion strategies operate at the pixel level of features, as shown in Fig. \ref{fig:fusion}.

For the fusion task of infrared image and visible light image, we believe that the two images are not obvious biased in feature selection. 
% Although the visible light image provides more details, it also needs semantic features. 
% Although infrared images have obvious semantic features, they also need outline and color low-level features. 
We decide to use the average strategy to obtain their fusion features, $F_{fuse}=(F_{vi}+F_{ir})/2$.

For the multi-focus image fusion tasks, as the focus of the image are different, the features of the focused area are more prominent than the unfocused area. 
We believe that the fused pixel should be the more obvious one. 
Therefore,we adopt the maximum value strategy, $F_{fuse}=\operatorname{max}(F_{vi}, F_{ir})$.

In addition to these two common fusion strategies, we propose a Softmax strategy, which can be used for multiple image fusion tasks at the same time. 
% This is an adaptive strategy. 
% If the corresponding pixels are both important, the two will account for the same proportion. 
% If one of the features pixel is weak, its proportion is small. 
To adaptively trade off the significance between the two input images, Softmax is employed to fuse the two features,
% The formula is as follows,  
$F_{fuse}=\operatorname{softmax}(F_{vi}, F_{ir})$.

\begin{figure*}[!ht]
	\centering
	\includegraphics[width=\linewidth]{./figures/irvifusion2}
	\caption{Comparison of our PPT Fusio with 18 state-of-the-art methods on one pair of Visible light and Infrared image in the (1) TNO dataset and (2) RoadScene dataset.}
	\label{fig:irvifusion}
\end{figure*}
\begin{table*}[!ht]
	\begin{center}
		\resizebox{\textwidth}{!}{
			\begin{tabular}{@{}|cc|cccccc|cccccc|@{}}
				\toprule
				\multicolumn{2}{|c|}{}      & \multicolumn{6}{c|}{TNO}                            & \multicolumn{6}{c|}{Road}                           \\ \cmidrule(l){3-14} 
				\multicolumn{2}{|c|}{\multirow{-2}{*}{Methods}} & SCD    & SSIM   & $FMI_{pixel} $               & $N_{abf}$   & $Q_S$     & CC     & SCD    & SSIM   & $FMI_{pixel} $               & $N_{abf}$   & $Q_S$     & CC     \\ \midrule
				\multicolumn{2}{|c|}{CBF}   & 1.3193     & 0.6142     & 0.8888     & 0.0621     & 0.7395     & 0.6560     & 0.8784     & 0.6028     & 0.8312     & 0.0388     & 0.7387     & 0.7118     \\
				\multicolumn{2}{|c|}{CVT}   & 1.5812     & 0.7025     & \textit{\textbf{0.9156}}  & 0.0275     & 0.8118     & 0.7111     & 1.3418     & 0.6641     & 0.8643     & 0.0318     & 0.7568     & 0.7439     \\
				\multicolumn{2}{|c|}{DTCWT} & 1.5829     & 0.7057     & {\color[HTML]{FE0000} \textit{\textbf{0.9186}}} & 0.0232     & 0.8182     & 0.7121     & 1.3329     & 0.6567     & 0.8509     & 0.0412     & 0.7326     & 0.7420     \\
				\multicolumn{2}{|c|}{GTF}   & 1.0516     & 0.6798     & 0.9056     & 0.0126     & 0.7536     & 0.6514     & 0.8072     & 0.6748     & 0.8654     & 0.0096     & 0.7037     & 0.6961     \\
				\multicolumn{2}{|c|}{MSVD}  & 1.5857     & 0.7360     & 0.9036     & \textit{\textbf{0.0022}}  & 0.7888     & 0.7280     & 1.3458     & 0.7128     & 0.8459     & 0.0034     & 0.7356     & 0.7518     \\
				\multicolumn{2}{|c|}{RP}    & 1.5769     & 0.6705     & 0.8929     & 0.0583     & 0.7542     & 0.7124     & 1.2829     & 0.6341     & 0.8408     & 0.0773     & 0.7229     & 0.7300     \\
				\multicolumn{2}{|c|}{DeepFuse}  & 1.5523     & 0.7135 & 0.9041     & 0.0202 & 0.7698     & 0.7243     & 0.5462     & 0.4601     & 0.8215     & 0.2213     & 0.5387     & 0.6687     \\
				\multicolumn{2}{|c|}{DenseFuse} & 1.5329     & 0.7108     & 0.9061     & 0.0352     & 0.7847     & 0.6966     & 1.3491     & 0.7404     & 0.8520     & {\color[HTML]{FE0000} \textit{\textbf{0.0001}}} & 0.7602     & 0.7543     \\
				\multicolumn{2}{|c|}{FusionGan} & 0.6876     & 0.6235     & 0.8875     & 0.0352     & 0.6422     & 0.6161     & 0.8671     & 0.6142     & 0.8398     & 0.0168     & 0.6433     & 0.7312     \\
				\multicolumn{2}{|c|}{IFCNN} & 1.6126     & 0.7168     & 0.9007     & 0.0346     & \textit{\textbf{0.8257}}  & 0.7004     & 1.3801     & 0.7046     & 0.8509     & 0.0315     & 0.7606     & {\color[HTML]{FE0000} \textit{\textbf{0.7647}}} \\
				\multicolumn{2}{|c|}{MdLatLRR}  & \textit{\textbf{1.6248}}  & 0.7306     & 0.9148     & 0.0216     & {\color[HTML]{FE0000} \textit{\textbf{0.8427}}} & 0.7137     & 1.3636     & 0.7369     & 0.8575     & \textit{\textbf{0.0004}}  & 0.7758     & 0.7527 \\
				\multicolumn{2}{|c|}{DDcGAN}    & 1.3831  & 0.5593     & 0.8764     & 0.1323     & 0.6571 & 0.7079     & 0.5462     & 0.4601     & 0.8215     & 0.2213     & 0.5387     & 0.6687     \\
				\multicolumn{2}{|c|}{ResNetFusion}             & 0.1937     & 0.4560     & 0.8692     & 0.0550     & 0.4247     & 0.3887     & 0.2179     & 0.3599     & 0.7963     & 0.0550 & 0.3462     & 0.5376     \\
				\multicolumn{2}{|c|}{Nestfuse}  & 1.5742     & 0.7057     & 0.9029     & 0.0428     & 0.7833     & 0.7006     & 1.2583     & 0.6666     & 0.8571     & 0.0459     & 0.6894     & 0.7539     \\
				\multicolumn{2}{|c|}{FusionDN}  & 1.6148     & 0.6201     & 0.8833     & 0.1540     & 0.7328     & 0.7170     & 1.1882     & 0.6454     & 0.8423     & 0.0780     & 0.7658     & 0.7204     \\
				\multicolumn{2}{|c|}{HybridMSD} & 1.5773     & 0.7094     & 0.9083     & 0.0435     & 0.8208     & 0.7072     & 1.2642     & 0.6961     & 0.8552     & 0.0460     & 0.7679     & \textit{\textbf{0.7565}}  \\
				\multicolumn{2}{|c|}{PMGI}  & 1.5738     & 0.6976     & 0.9001     & 0.0340     & 0.7632     & 0.7281     & 1.0989     & 0.6640     & 0.8487     & 0.0146     & 0.7606     & 0.7195  \\
				\multicolumn{2}{|c|}{U2Fusion}  & 1.5946     & 0.6758     & 0.8942     & 0.0800     & 0.7439     & 0.7238     & 1.3551     & 0.6813     & 0.8453     & 0.0671     & 0.7831     & 0.7266     \\
				& Average        & {\color[HTML]{FE0000} \textit{\textbf{1.6261}}} & \textit{\textbf{0.7487}}  & 0.8954     & 0.0060     & 0.7719     & \textit{\textbf{0.7313}}  & {\color[HTML]{FE0000} \textit{\textbf{1.5888}}} & {\color[HTML]{FE0000} \textit{\textbf{0.7528}}} & {\color[HTML]{FE0000} \textit{\textbf{0.8982}}} & 0.0109     & {\color[HTML]{FE0000} \textit{\textbf{0.7953}}} & 0.7183     \\
				\multirow{-2}{*}{Ours}        & Softmax        & 1.5858 & {\color[HTML]{FE0000} \textit{\textbf{0.7568}}} & 0.9077     & {\color[HTML]{FE0000} \textit{\textbf{0.0005}}} & 0.7945     & {\color[HTML]{FE0000} \textit{\textbf{0.7401}}} & \textit{\textbf{1.5870}} & \textit{\textbf{0.7505}} & \textit{\textbf{0.8959}} & 0.0145     & \textit{\textbf{0.7836}} & 0.7159     \\ \bottomrule
			\end{tabular}
		}
	\end{center}
	\caption{Visible and Infrared Image Fusion Quantitative Analysis. This table contains quantitative analysis indicators of the TNO and RoadScene datasets.}
	\label{table:irvi}
\end{table*}
\begin{table*}[!htp]
	\begin{center}
		\resizebox{\textwidth}{!}{
			\begin{tabular}{@{}|cl|cccccccccc|@{}}
				\toprule
				\multicolumn{2}{|c|}{Methods}   & SD      & EN     & $Q_S$   & CC     & VIFF   & CrossEntropy              & SSIM   & $FMI_{pixel}$                & MI      &  $N_{abf}$   \\ \midrule
				\multicolumn{2}{|c|}{GFF}       & 50.1114 & 7.2605 & 0.9167 & 0.9545 & 0.7699 & 0.0634 & 0.8147 & 0.8867 & 14.5211 & 0.0115 \\
				\multicolumn{2}{|c|}{LPSR}      & 50.6157 & 7.2640 & {\color[HTML]{FE0000} \textit{\textbf{0.9185}}} & 0.9557 & {\color[HTML]{FE0000} \textit{\textbf{0.8102}}} & 0.0672 & 0.8154 & 0.8845 & 14.5279 & 0.0731 \\
				\multicolumn{2}{|c|}{MFCNN}     & 50.2571 & 7.2538 & 0.9127 & 0.9544 & 0.7699 & 0.0643 & 0.8134 & 0.8870 & 14.5075 & {\color[HTML]{FE0000} \textit{\textbf{0.0015}}} \\
				\multicolumn{2}{|c|}{densefuse} & 52.6174 & 7.2882 & 0.8875 & 0.9671 & 0.7781 & 0.5948 & 0.8382 & 0.8744 & 14.5765 & 0.0110 \\
				\multicolumn{2}{|c|}{IFCNN}     & 49.7551 & 7.2384 & 0.9112 & 0.9600 & 0.7742 & 0.0779 & 0.8300 & 0.8735 & 14.4768 & 0.0762 \\
				& Max  & {\color[HTML]{FE0000} \textit{\textbf{60.6446}}} & \textit{\textbf{7.5239}}  & 0.9176 & \textit{\textbf{0.9870}}  & \textit{\textbf{0.8086}}  & {\color[HTML]{FE0000} \textit{\textbf{0.0242}}} & \textit{\textbf{0.8802}}  & \textit{\textbf{0.8911}}  & \textit{\textbf{15.0478}}  & \textit{\textbf{0.0101}}  \\
				\multirow{-2}{*}{Ours}   & Softmax   & \textit{\textbf{60.3025}}  & {\color[HTML]{FE0000} \textit{\textbf{7.5252}}} & \textit{\textbf{0.9182}}  & {\color[HTML]{FE0000} \textit{\textbf{0.9871}}} & 0.8020 & \textit{\textbf{0.0278}}  & {\color[HTML]{FE0000} \textit{\textbf{0.8818}}} & {\color[HTML]{FE0000} \textit{\textbf{0.8911}}} & {\color[HTML]{FE0000} \textit{\textbf{15.0504}}} & 0.0103 \\ \bottomrule
			\end{tabular}
		}
	\end{center}
	\caption{Multi-focus Image Fusion Quantitative Analysis. This table contains quantitative analysis indicators of the Lytro datasets.}
	\label{table:focus}
\end{table*}

\section{Evaluation}
\subsection{Datasets and Implementation}
PPT is designed as a general image feature extraction model for low-level vision tasks. 
We train this model using MS-COCO-2014 \citep{lin2014microsoft}, which contains 82612 images. 
PPT focuses more on extracting features and reconstructing images. 
PPT does not have to recognize very abstract semantic features. 
Though Imagenet has sufficient image texture and rich object recognition, the amount of COCO is sufficient for PPT training to expectations, without the need for a huge data set like other Transformer models.

After the network training, we do not need to adjust the PPT network structure, nor do we need to retrain the weights for subsequent tasks. 
We perform test experiments on four image fusion tasks, \textit{i.e.}, 1) infrared and visible light image fusion, 2)  multi-focus image fusion, 3)  medical image fusion and 4) multi-focus image fusion.

In the infrared image and visible light image fusion task, we use the TNO dataset \citep{toet2014tno} and the RoadScene dataset \citep{xu2020fusiondn}. 
For the RoadScene dataset, we convert the images to gray scale to keep the the visible light image channels consistent with infrared image. 
For the multi-focus image fusion task, we use the Lytro dataset \citep{Nejati2015Multi}. 
The Lytro image are split according to the RGB channels to obtain three pairs of images. 
The fusion result is merged according to the RGB to obtain a fused image.
For the medical image fusion task, we use the Harvard dataset\citep{Harvard}. 
For the multi-focus image fusion task, we use the dataset in  \citep{Cai2018medataset}. 

As the network input is a fixed size $W\times H$, we split the input image into several patches with a sliding window of $W\times H$ size, filling the insufficient area with the value 128 (the pixel range is 0$\sim$255). 
After fusing each patch pair, the final fused image is obtained by splicing according to the order of patch split.

\subsection{Experiments Setting}
We input the $256\times256$ image to the network, and the size of the patch $p=32$.
The optimizer is selected as Adam \citep{kingma2014adam} with a learning rate 1e-4. The batch size is 1. 
We set the total training times of the network to 50 times. 
The experiments are performed on an NVIDIA Geforce GTX1080 GPU and 3.60GHz Intel Core i7-6850K with 64GB of memory.
PPT uses fewer computing resources, it can use 8GB GPU memory to process 512-size images, which is parsimonious than classical Transformers. 

\subsection{Qualitative and Quantitative Analysis}
%\xu{(The analysis in this subsection is very confusing. Except the numerical advantages, more underlying insight should be involved when analysing the experimental results.)}
We conduct qualitative and quantitative experiments to vali-date the effectivenes.
We apply the PPT module to image fusion, compare with the stat of the art methods, and use relevant indicators to analyze the quality of image fusion.

% \emph{\fu{(I moved the list of comparison methods and indicators to the first paragraph of this section. Reduce the space below to avoid redundancy.)}}
We compare PPT Fusion with state-of-the-a methods contain traditional, auto-encoder based and end-to-end based methods,
such as 
Cross Bilateral Filter fusion method (CBF) \citep{kumar2015image}, 
Curvelet Transform (CVT) \citep{nencini2007remote}, 
Dual-Tree Complex Wavelet Transform (DTCWT) \citep{lewis2007pixel}, 
Gradient Transfer(GTF) \citep{ma2016infrared}, 
Multi-resolution Singular Value Decomposition (MSVD) \citep{naidu2011image}, 
Ratio of Low-pass Pyramid (RP) \citep{toet1989image}, 
Deepfuse \citep{prabhakar2017deepfuse}, DenseFuse \citep{li2018densefuse}, 
Guided Filtering Fusion (GFF) \citep{li2013image},
Laplace Pyramid Sparse Representation (LPSR) \citep{liu2015general},
MFCNN \citep{liu2017multi}, 
IFCNN \citep{zhang2020ifcnn},
FusionGan \citep{ma2019fusiongan}, 
MDLatLRR \citep{li2020mdlatlrr}, 
DDcGan  \citep{ma2020ddcgan:}, 
ResNetFusion \citep{ma2020infrared}, 
NestFuse \citep{li2020nestfuse}, 
FusionDN \citep{xu2020fusiondn}, 
HybridMSD \citep{zhou2016perceptual}, 
PMGI \citep{zhang2020rethinking},  
multi-exposure image fusion \citep{li2020fmmef},
Guided Filter focus region detection (GFDF)\citep{QIU2019GFDF} and 
U2Fusion \citep{xu2020u2fusion} respectively.

We also exploit related indicators to quantitatively evaluate the fusion image quality, with different appropriate indicators being used for different tasks,
namely Sum of Correlation Coefficients (SCD) \citep{aslantas2015new}, Structural SIMilarity (SSIM) \citep{wang2004image}, pixel feature mutual information($FMI_{pixel}$) \citep{haghighat2014fast}, $N_{abf}$ \citep{kumar2013multifocus}, $Q_S$ \citep{liu2011objective}, correlation coefficient (CC) \citep{han2008study}, Entropy(EN) \citep{roberts2008assessment}, Visual Information Fidelity(VIFF) \citep{han2013a}, 
Cross Entropy \citep{kumar2013multifocus}, Mutual Information (MI)\citep{peng2005feature},  
Fast Mutual Information ($FMI_{w}$ and $FMI_{dct}$) \citep{haghighat2014fast},
modified structural similarity (MS-SSIM)\citep{ma2015perceptual},$N_{abf}$ \citep{kumar2013multifocus}, Visual Information Fidelity (VIF) \citep{sheikh2006image},Standard Deviation of Image (SD)\citep{rao1997fibre},
Revised Mutual Information (QMI)\citep{cvejic2006image}, Nonlinear Correlation Information Entropy (QNCIE)\citep{inbook} and Phase Congruency Measurement (QP)\citep{Zhao2006Performance}.

In particular, SCD and CC calculates the correlation coefficients between images. 
SSIM and $Q_S$ calculate the similarity between images.
$FMI_{pixel}$ calculate the mutual information between features. 
$N_{abf}$ represents the ratio of noise added to the final image. 
EN measure the amount of information. 
VIFF is used to measure the loss of image information to the distortion process. 
Cross Entropy and MI measure the degree of information correlation between images.
Among them, the lower the value of the $N^{ab/f}$ and the Cross Entropy and the higher other values, the better the fusion quality of the approach.

\subsubsection{Visible and Infrared Image Fusion}

As shown in Fig. \ref{fig:irvifusion}, we report the results of all approaches and highlight some specific local areas. 
It can be seen that the fusion result of our PPT Fusion retains the necessary person radiation information. 
The global semantic feature of our results is more obvious, that is, the contrast between the sky and the house.
After highlighting the details of the branches, our results reflect more details from both the visible light image and the infrared image.

%We using six related indicators to quantitatively evaluate the fusion quality, namely Sum of Correlation Coefficients (SCD) \citep{aslantas2015new}, Structural SIMilarity (SSIM) \citep{wang2004image}, pixel feature mutual information($FMI_{pixel}$) \citep{haghighat2014fast}, $N_{abf}$ \citep{kumar2013multifocus}, $Q_S$ \citep{liu2011objective} and correlation coefficient (CC) \citep{han2008study}.
%SCD and CC calculates the correlation coefficients between images. 
%SSIM and $Q_S$ calculate the similarity between images.
%$FMI_{pixel}$ calculate the mutual information between features. 
%$N_{abf}$ represents the ratio of noise added to the final image. 
%Among them, the lower the value of the $N^{ab/f}$, and the higher other values, the better the fusion quality of the approach.

As shown in Table. \ref{table:irvi}, the best value in the quality table is made the bold red font in italic, and the second-best value is in the bold black font in italic. 
It can be seen that PPT Fusion rank in top 2 in multiple indicators. 
Other indicators are also better than most methods. 
It can be demonstrated that PPT Fusion maintains a effective structural similarity with the source images, preserving a large information correlation with the source images, without introducing noise, artifacts, etc.

\subsubsection{Multi-focus Image Fusion}
%We compare PPT Fusion with the five state-of-the-art methods, and they are Guided Filtering Fusion(GFF) \citep{li2013image},
%Laplace Pyramid Sparse Representation(LPSR) \citep{liu2015general},
%MFCNN \citep{liu2017multi}, DenseFuse \citep{li2018densefuse} and IFCNN \citep{zhang2020ifcnn}, as shown in Fig. \ref{fig:lytro}.

\begin{figure}[!ht]
	\centering
	\includegraphics[width=7cm]{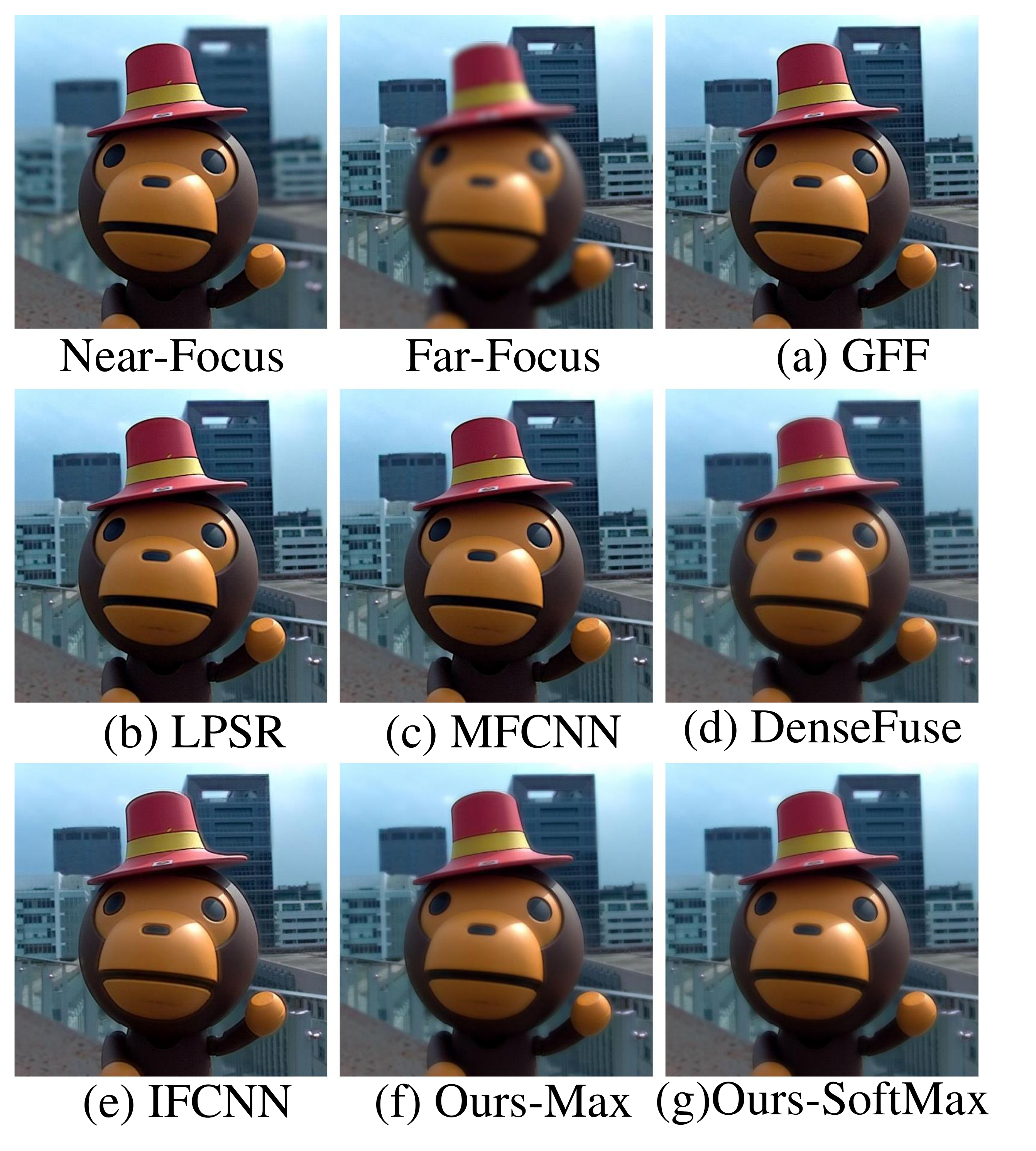}
	\caption{Comparison of PPT Fusion result with five state-of-the-art methods on one pair of multi-focus  image in the Lytro dataset.}
	\label{fig:lytro}
\end{figure}

As shown in Fig. \ref{fig:lytro}, we report the results of all approaches and highlight some specific local areas.

%On the basis of the previous five indicators: SCD, SSIM, $FMI_{pixel}$, $N_{abf}$, $Q_S$ and CC, we added four additional indicators, namely Entropy(EN) \citep{roberts2008assessment}, Visual Information Fidelity(VIFF) \citep{han2013a}, 
%Cross Entropy \citep{kumar2013multifocus}, Mutual Information (MI)\citep{peng2005feature}.
%EN measure the amount of information.
%VIFF is used to measure the loss of image information to the distortion process. CrossEntropy and MI measure the degree of information correlation between images.
%Among them, the lower the value of the CrossEntropy and the higher other values, the better the fusion quality of the approach.

From Table \ref{table:focus}, we can see that PPT Fusion can rank in top 2 in all indicators. 
This shows that the fusion image of PPT Fusion effectively extract the source details while making the generated image clear enough.

\begin{figure*}[!h]
	\centering
	\includegraphics[width=14cm]{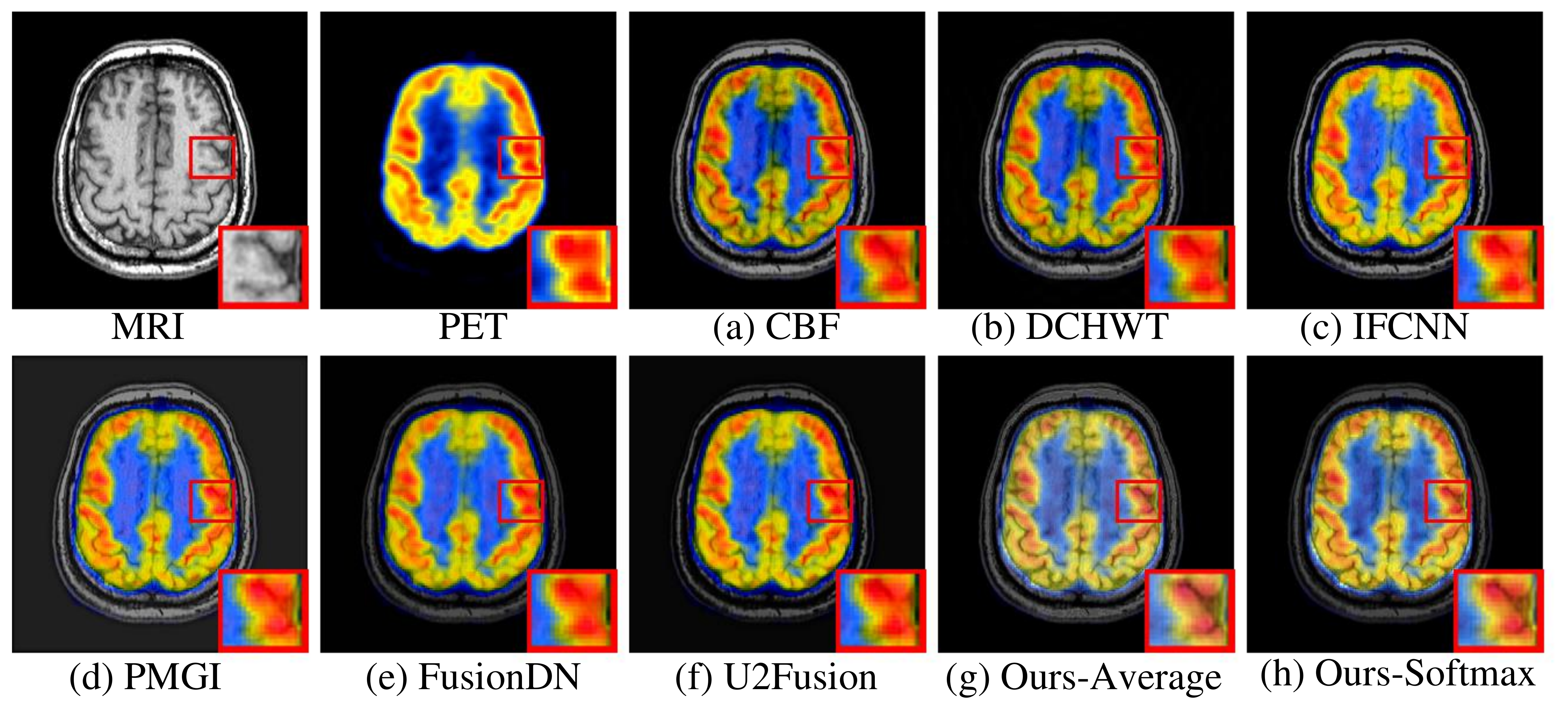}
	\caption{Comparison of PPT Fusion result with 6 state-of-the-art methods on one pair of medical image in the Harvard Medical Image dataset.}
	\label{fig:harvard}
\end{figure*}

\subsubsection{Medical Image Fusion}
%We compare PPT Fusion with the six state-of-the-art methods, including Cross Bilateral Filter fusion method(CBF) \citep{kumar2015image}, Dual-Tree Complex Wavelet Transform(DTCWT) \citep{lewis2007pixel}, IFCNN \citep{zhang2020ifcnn}, PMGI \citep{zhang2020rethinking}, FusionDN \citep{xu2020fusiondn} and U2Fusion \citep{xu2020u2fusion}, respectively.

As shown in Fig. \ref{fig:harvard}, we report the results of all approaches and highlight some specific local areas. 
The texture structure information of MRI and the color distribution information of PET are both very important. It can be seen that our results not only retain the color distribution, but also significantly retain the texture information. The results of other methods lose a lot of MRI information.

\begin{table*}[!htp]
	\begin{center}
		\resizebox{\textwidth}{!}{
			\begin{tabular}{@{}ccccccccccccccccc@{}}
				\toprule
				\multicolumn{2}{c}{Methods} & SCD & $FMI_w$ & $FMI_dct$ & SSIM & MS-SSIM & $N_{abf}$ & VIF & SD & $Q_{MI}$ & $Q_{S}$ & $Q_{NCIE}$ & MI & CC & VIFF & $Q_{P}$ \\ \midrule
				\multicolumn{2}{c}{PMGI} & 0.4087 & 0.1778 & 0.1468 & 0.1396 & 0.5801 & 0.0249 & 0.2634 & 52.7224 & 0.4978 & 0.1678 & 0.8048 & 2.2892 & 0.8613 & 0.1660 & 0.0574 \\
				\multicolumn{2}{c}{U2Fusion} & 0.4754 & 0.2154 & {\color[HTML]{FF0000} \textit{\textbf{0.1712}}} & 0.2005 & 0.5696 & \textit{\textbf{0.0191}} & 0.3018 & 57.6757 & 0.5786 & 0.2102 & 0.8052 & 2.4827 & 0.8718 & 0.1840 & {\color[HTML]{FF0000} \textit{\textbf{0.0646}}} \\
				\multicolumn{2}{c}{IFCNN} & 0.4887 & 0.2149 & 0.1424 & 0.5780 & 0.6090 & 0.0309 & {\color[HTML]{FF0000} \textit{\textbf{0.3351}}} & 64.2122 & 0.5735 & 0.5399 & 0.8051 & 2.4056 & 0.8625 & 0.2029 & 0.0602 \\
				\multicolumn{2}{c}{DCHWT} & 0.4187 & 0.1607 & 0.1486 & 0.5531 & 0.6085 & 0.0222 & 0.2919 & 56.6969 & 0.4911 & 0.5243 & 0.8046 & 2.2234 & 0.8634 & 0.1878 & 0.0624 \\
				\multicolumn{2}{c}{CSR} & 0.4219 & 0.2071 & 0.1386 & 0.5629 & 0.5902 & 0.0299 & 0.2871 & 55.2012 & 0.5144 & 0.5309 & 0.8046 & 2.2151 & 0.8529 & 0.1731 & 0.0489 \\
				\multicolumn{2}{c}{CBF} & 0.4459 & 0.2157 & 0.1472 & 0.5695 & 0.6012 & 0.0291 & 0.2983 & 56.2595 & 0.5471 & 0.5374 & 0.8048 & 2.2932 & 0.8574 & 0.1836 & 0.0536 \\
				& Average & \textit{\textbf{0.9100}} & \textit{\textbf{0.2180}} & 0.1461 & \textit{\textbf{0.5833}} & \textit{\textbf{0.6106}} & 0.0196 & 0.2936 & {\color[HTML]{FF0000} \textit{\textbf{67.2116}}} & \textit{\textbf{0.6007}} & \textit{\textbf{0.5445}} & \textit{\textbf{0.8054}} & \textit{\textbf{2.5538}} & \textit{\textbf{0.8833}} & {\color[HTML]{FF0000} \textit{\textbf{0.2130}}} & 0.0519 \\
				\multirow{-2}{*}{Ours} & Max & {\color[HTML]{FF0000} \textit{\textbf{0.9661}}} & {\color[HTML]{FF0000} \textit{\textbf{0.2219}}} & \textit{\textbf{0.1705}} & {\color[HTML]{FF0000} \textit{\textbf{0.5924}}} & {\color[HTML]{FF0000} \textit{\textbf{0.6134}}} & {\color[HTML]{FF0000} \textit{\textbf{0.0187}}} & \textit{\textbf{0.3236}} & \textit{\textbf{66.5805}} & {\color[HTML]{FF0000} \textit{\textbf{0.6133}}} & {\color[HTML]{FF0000} \textit{\textbf{0.5476}}} & {\color[HTML]{FF0000} \textit{\textbf{0.8055}}} & {\color[HTML]{FF0000} \textit{\textbf{2.6003}}} & {\color[HTML]{FF0000} \textit{\textbf{0.8875}}} & \textit{\textbf{0.2129}} & \textit{\textbf{0.0633}} \\ \bottomrule
			\end{tabular}
		}
	\end{center}
	\caption{Medical Image Fusion Quantitative Analysis. This table contains quantitative analysis indicators of the Harvard Medical Image datasets.}
	\label{table:medical}
\end{table*}

%We using fifteen related indicators to quantitatively evaluate the fusion quality, namely Sum of Correlation Coefficients (SCD) \citep{aslantas2015new}, 
%Fast Mutual Information ($FMI_{w}$ and $FMI_{dct}$) \citep{haghighat2014fast},
%Structural Similarity (SSIM) \citep{wang2004image}, modified structural similarity (MS-SSIM)\citep{ma2015perceptual},$N_{abf}$ \citep{kumar2013multifocus}, Visual Information Fidelity (VIF) \citep{sheikh2006image},Standard Deviation of Image (SD)\citep{rao1997fibre},
%Revised Mutual Information (QMI)\citep{cvejic2006image}, $Q_S$ \citep{liu2011objective}, Nonlinear Correlation Information Entropy (QNCIE)\citep{inbook}, correlation coefficient (CC) \citep{han2008study} and Phase Congruency Measurement (QP)\citep{Zhao2006Performance}.

As shown in Table. \ref{table:medical}, the best value in the quality table is made the bold red font in italic, and the second-best value is in the bold black font in italic. 
It can be seen that PPT Fusion rank in top 2 in all indicators. 
Other indicators are also better than most methods.

\subsubsection{Multi-exposure Image Fusion}
%We compare PPT Fusion with the six state-of-the-art methods,
%including Fast multi-scale structural patch decomposition for multi-exposure image fusion \citep{li2020fmmef},
%Guided Filter focus region detection(GFDF)\citep{QIU2019GFDF},
%FusionDN \citep{xu2020fusiondn}, 
%IFCNN \citep{zhang2020ifcnn}, 
%PMGI \citep{zhang2020rethinking}, and U2Fusion \citep{xu2020u2fusion}, respectively.

\begin{figure*}[!ht]
	\centering
	\includegraphics[width=\linewidth]{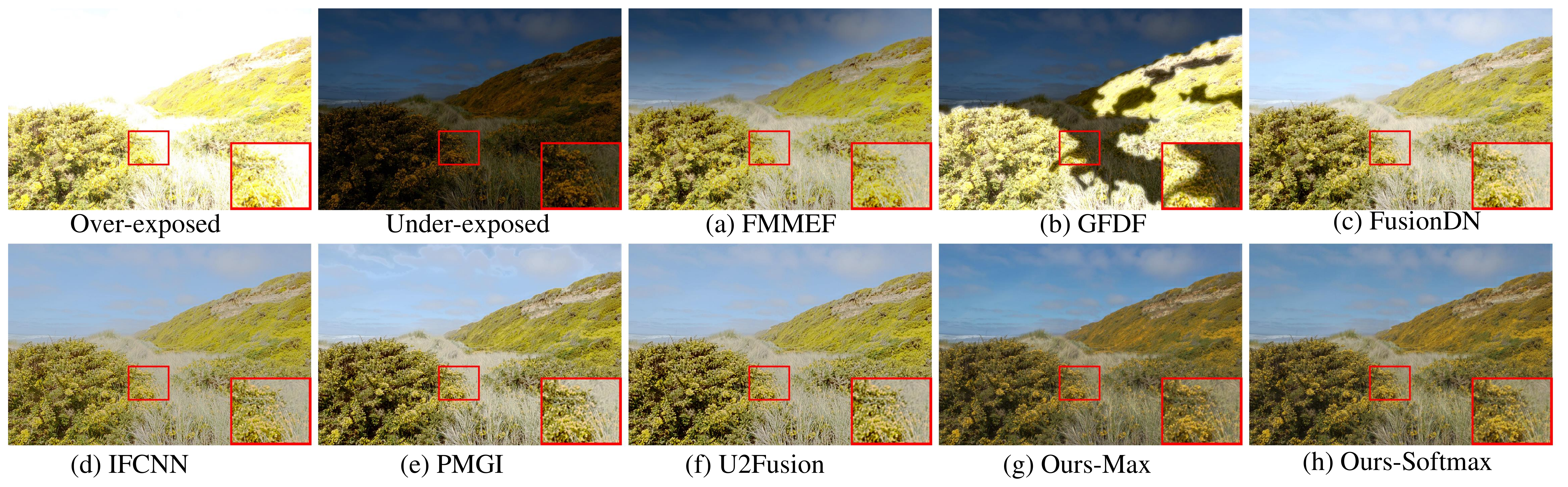}
	\caption{Comparison of PPT Fusion result with 6 state-of-the-art methods on one pair of multi-exposure image in the HDR dataset.}
	\label{fig:exposure}
\end{figure*}

Both over-exposed and under-exposed information are incorrect. Our results do not look as bright as other methods, but they actually retain more detailed information. For example, the contrast between flowers and trees, the texture of weeds, and the distribution and shape of clouds in the sky. Other methods tend to retain the over-exposed information, resulting in unclear details.

%We using fifteen related indicators to quantitatively evaluate the fusion quality, namely Sum of Correlation Coefficients (SCD) \citep{aslantas2015new}, 
%Fast Mutual Information ($FMI_{w}$ and $FMI_{dct}$) \citep{haghighat2014fast},
%Structural Similarity (SSIM) \citep{wang2004image}, modified structural similarity (MS-SSIM)\citep{ma2015perceptual},$N_{abf}$ \citep{kumar2013multifocus}, Visual Information Fidelity (VIF) \citep{sheikh2006image},Standard Deviation of Image (SD)\citep{rao1997fibre},
%Revised Mutual Information (QMI)\citep{cvejic2006image}, $Q_S$ \citep{liu2011objective}, Nonlinear Correlation Information Entropy (QNCIE)\citep{inbook}, correlation coefficient (CC) \citep{han2008study} and Phase Congruency Measurement (QP)\citep{Zhao2006Performance}.

\begin{table}[!htp]
	\begin{center}
		\resizebox{\linewidth}{!}{
			\begin{tabular}{@{}cccccccc@{}}
				\toprule
				\multicolumn{2}{c}{Methods} & $N_{abf}$ & $Q_{MI}$ & $Q_{NCIE}$ & $MI$ & $Q_{P}$ & $Q_{CB}$ \\ \midrule
				\multicolumn{2}{c}{U2Fusion} & 0.1107 & 0.7651 & 0.8188 & 4.9059 & 0.6950 & 0.4505 \\
				\multicolumn{2}{c}{GFDF} & 0.0476 & 0.9034 & 0.8276 & 6.0607 & 0.5276 & 0.4645 \\
				\multicolumn{2}{c}{PMGI} & 0.1076 & 0.8772 & 0.8232 & 5.7564 & 0.0000 & 0.4597 \\
				\multicolumn{2}{c}{IFCNN} & 0.0585 & 0.7714 & 0.8188 & 4.9013 & 0.7147 & 0.3740 \\
				\multicolumn{2}{c}{DeepFuse} & 0.0371 & 1.0276 & 0.8125 & {\color[HTML]{FF0000} \textit{\textbf{7.1086}}} & {\color[HTML]{FF0000} \textit{\textbf{0.7648}}} & 0.4122 \\
				\multicolumn{2}{c}{FMMEF} & 0.0212 & 0.4972 & 0.8143 & 3.3467 & 0.0000 & 0.4573 \\
				\multicolumn{2}{c}{GFF} & 0.0290 & 0.7246 & 0.8236 & 4.9657 & 0.0000 & 0.4749 \\
				& Max & {\color[HTML]{FF0000} \textit{\textbf{0.0162}}} & \textit{\textbf{1.0355}} & \textit{\textbf{0.8289}} & 6.6366 & 0.7161 & {\color[HTML]{FF0000} \textit{\textbf{0.4829}}} \\
				\multirow{-2}{*}{Ours} & Softmax & \textit{\textbf{0.0203}} & {\color[HTML]{FF0000} \textit{\textbf{1.0360}}} & {\color[HTML]{FF0000} \textit{\textbf{0.8290}}} & \textit{\textbf{6.6598}} & \textit{\textbf{0.7190}} & \textit{\textbf{0.4784}} \\ \bottomrule
			\end{tabular}
		}
	\end{center}
	\caption{Multi-exposure Image Fusion Quantitative Analysis. This table contains quantitative analysis indicators of the Multi-exposure Image datasets.}
	\label{table:exposure}
\end{table}

As shown in Table. \ref{table:exposure}, the best value in the quality table is made the bold red font in italic, and the second-best value is in the bold black font in italic. 
It can be seen that PPT Fusion rank in top 2 in all indicators. 
Other indicators are also better than most methods. 

\subsection{Results Analysis}
% \emph{\fu{("more underlying insight". Analyze the advantages, and then analyze the network structure. Write it like this, I guess?)}}
In the subjective comparison of fused images, it can be clearly seen that the PPT fusion results exhibit more contrast information, such as TNO infrared visible light images, medical images and multi-exposure images. 
The self-attention in Patch Transformer combined with Pyramid Transformer can obtain the context information of each pixel of the image from the long-range perception. 
This helps to maintain the consistency of the same semantic object, and to distinguish the foreground object from the background significantly.
At the same time, just like convolution, Patch Transformer with minimum receptive field focuses on extracting the texture structure of the image. 
So that the PPT fusion results preserve sufficient detailed information. 
The details of the object are more clear in the obvious contrast environment.
Quantitative indicators of our method show that our PPT results contain more information, texture details, stronger contrast, and a higher similarity with the source image with less distortion or artifacts, verifying the effectiveness of the PPT network structure.

% \fu{\emph{("the merit of this structure, \textit{i.e.}, speed, generalisation, robustness, training difficulty.")}}

After training on about 80,000 pictures with 10GB memory, PPT model can be used for several different image fusion tasks. 
The training difficulty is much lower than other Transformer model. 
PPT can be used without retraining, it has strong generalization. 
Although our PPT model cannot get the first place in all tasks. 
PPT can decompose image features well and perform feature fusion in the feature space. 
We think PPT feature extraction ability can be used successfully in other low-level vision tasks.

%\xu{(As there is no ablation study for the PPT network. It is recommended to add another subsection to present the merit of this structure, \textit{i.e.}, speed, generalisation, robustness, training difficulty.)}

\section{Conclusions}
In this study, we propose a feature extraction module that uses Fully-Transformer, termed as the Pyramid Patch Transformer (PPT) module. 
%First, the Patch Transformer we proposed can map  high-resolution images to feature space without resolution loss.
First, the Patch Transformer we proposed can not only perceive the local features of the image, but also perceive non-local context information.
Second, we propose the Pyramid Transformer with transformer receptive field to extract 
%local information and global information 
texture information and context information from images. 
The PPT module can map images into a set of multi-scale, multi-dimensional, and multi-angle features. 
We successfully apply the PPT module to different image fusion tasks and achieve the state-of-the-art. 
This proves that using a Fully-Transformer and designing a reasonably structure can represent the image features %without loss information
with its extracted local features and global features, demonstrating the effectiveness and universality of the PPT module. 
We use image fusion to verify the feasibility of PPT in the low-level vision applications. 
However, high-level visual tasks such as classification, detection, segmentation are not considered. Just as ViT can only handle advanced semantic visual tasks, PPT is typically designed for low-level visual tasks.
We believe that the propsoed PPT module has reference significance for low-level vision tasks and image generation tasks.



%%       \citet{<label>} ==> Jones et al. [21]
%%       \citep{<label>} ==> [21]
%%

\bibliographystyle{elsarticle-num-names} 

\bibliography{paper}


%\begin{thebibliography}{00}
%
%%% \bibitem[Author(year)]{label}
%%% Text of bibliographic item
%\end{thebibliography}

\end{document}